\documentclass{article}
\usepackage[accepted]{icml2017}
\usepackage{times}
\usepackage{hyperref}
\usepackage{url}
\usepackage[fleqn,tbtags]{amsmath}
\usepackage{amsfonts,amssymb}
\usepackage{graphicx}
\usepackage{subfigure}
\usepackage{natbib}
\usepackage{bm}
\usepackage{subfloat}
\usepackage{multirow}
\usepackage{colortbl}
\usepackage{hyperref}
\usepackage{enumitem}
\graphicspath{{figures/}}

\newcommand{\tens}[1]{{\mathcal{#1}}}
\newcommand{\reals}{\mathbb{R}}
\newcommand{\conv}{{\star}}

\newcommand{\btheta}{\bm{\theta}}
\newcommand{\bsigma}{\bm{\sigma}}

\iftrue
\newcommand{\jpark}[1]{{\color{blue}[\textbf{\sc Jongsoo}: \textit{#1}]}}
\newcommand{\ptang}[1]{{\color{green}[\textbf{\sc }: \textit{#1}]}}
\newcommand{\sli}[1]{{\color{red}[\textbf{\sc Sheng}: \textit{#1}]}}
\else
\newcommand{\jpark}[1]{}
\newcommand{\ptang}[1]{}
\newcommand{\sli}[1]{}
\fi

\icmltitlerunning{Enabling Sparse Winograd Convolution by Native Pruning}

\begin{document}

\twocolumn[\icmltitle{
Enabling Sparse Winograd Convolution by Native Pruning
}

\icmlsetsymbol{equal}{*}
\icmlsetsymbol{old}{$\dagger$}

\begin{icmlauthorlist}
\icmlauthor{Sheng Li}{equal,google,old}
\icmlauthor{Jongsoo Park}{equal,facebook,old}
\icmlauthor{Ping Tak Peter Tang}{equal,intel}
\end{icmlauthorlist}

\icmlaffiliation{intel}{Intel Labs}
\icmlaffiliation{google}{Google}
\icmlaffiliation{facebook}{Facebook}

\icmlcorrespondingauthor{Sheng Li}{lsheng@google.com}
\icmlcorrespondingauthor{Jongsoo Park}{jongsoo@fb.com}
\icmlcorrespondingauthor{Ping Tak Peter Tang}{peter.tang@intel.com}

\icmlkeywords{CNN, Winograd, Sparse, Pruning, Speedup}

\vskip 0.3in
]
\printAffiliationsAndNotice{\icmlEqualContribution}





\begin{abstract}

Sparse methods and the use of Winograd convolutions are two orthogonal approaches,
each of which significantly accelerates convolution computations in modern CNNs.
Sparse Winograd merges these two and thus has the potential to offer a combined
performance benefit. Nevertheless, training convolution layers so that the resulting
Winograd kernels are sparse has not hitherto been very successful. 
By introducing a Winograd layer in place of a standard convolution layer,
we can learn and prune Winograd coefficients ``natively'' and obtain 
sparsity level beyond 90\% with only 0.1\% accuracy loss with
AlexNet on ImageNet dataset.
Furthermore, we present a sparse Winograd convolution algorithm and implementation that exploits
the sparsity, achieving up to 31.7 effective TFLOP/s in 32-bit precision on a
latest Intel Xeon CPU, which corresponds to a 5.4$\times$ speedup over a
state-of-the-art dense convolution implementation.
\end{abstract}

\vspace{0.2cm}
\section{Introduction}\label{sec:introduction}

\begin{figure*}[t]
\includegraphics[width=\textwidth]{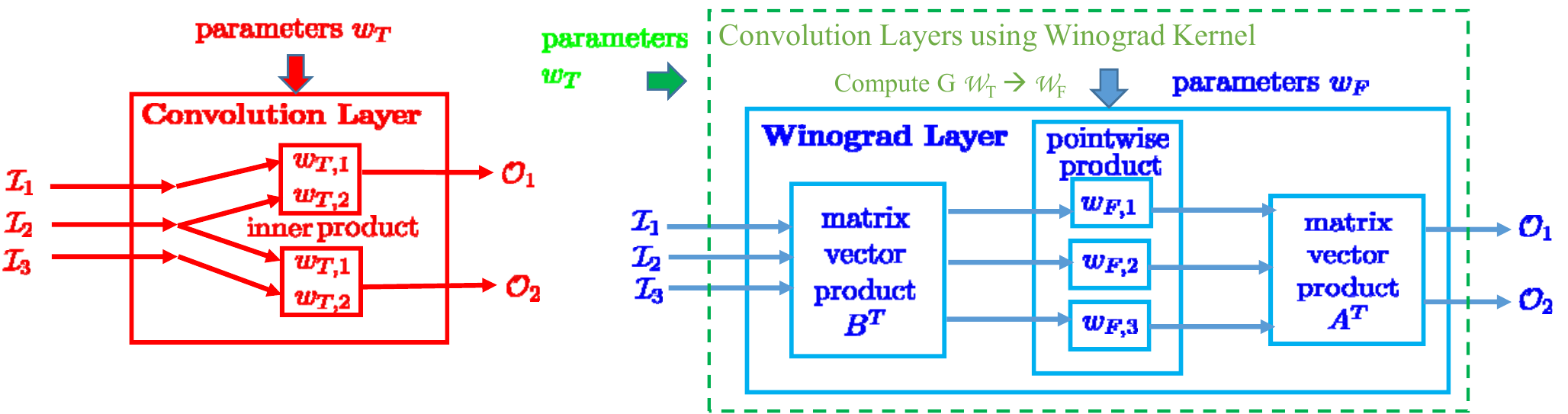}%
\caption{\small
\small
Schematic illustration for 1-D convolution and 1-D Winograd Convolution. The red schematic on the left illustrates a traditional
convolution layer. This example takes the inputs ${\cal I}_j$, $j=1,2,3$,
and kernel parameters $w_{T,j}$, $j=1,2$, and produces the output
${\cal O}_j$, $j=1,2$.
The blue schematic on the right illustrates a corresponding
Winograd layer. Note that it takes three independent kernel
inputs $w_{F,j}$, $j=1,2,3$. Given any inputs ${\cal I}_j$,
the Winograd layer produces two outputs ${\cal O}_j$.
The standard Winograd-as-a-kernel implements a convolution layer
using the Winograd computation method, and is illustrated by the
green dotted line schematic. It takes the spatial convolution
parameters (two coefficients) $w_T$ and the input ${\cal I}$.
Internally, it applies the (non-invertible) transform to $w_T$
that yields $w_F = G w_T$, which is then used in the Winograd
computation. Consequently, the set of $w_F$ used here is at most
two dimensional, instead of the Winograd layer that can exploit
a three-dimensional input space for $w_F$.
}
\label{fig:architecture-explained}
\vspace{-0.2cm}
\end{figure*}

Convolution neural networks (CNN) have achieved undisputed success in many practical applications.
These deep neural networks typically contain multiple layers,
many (though not all) of which perform the namesake computation of convolution.
A convolution layer is an architecture whose connection between
an input and output tensor is via a number of convolution kernels,
and the basic arithmetic operations are that of multiply-accumulate.
Because over 90\%
of the computation during inference and
training of recent CNN designs is in convolutions~\citep{alexnet, googlenet}, different strategies have been devised to speed up this core operation.
Sparse methods is one such strategy. Here, many of the
convolution kernel coefficients are made zero by some pruning
or compression techniques \citep{Brain_Damage:LeCun,scnn_lebedev,scnn}.
Sparsity exploiting convolution implementations are then employed in the actual
inference process. \cite{iclr17} reported a three-fold
speed up on the convolution layers when an optimized sparse
convolution implementation is applied on well pruned kernels.

Orthogonal to sparse methods, transform
methods such as FFT~\citep{fft-cnn,fbfft} or
Winograd transformation~\citep{winograd1980arithmetic,winograd} have proved to be successful as well.
For the typical small convolution sizes (e.g., 3$\times$3) that arise in CNNs, the Winograd-kernel approach
is more effective than FFT and has demonstrated more than twofold speed up
over well-implemented spatial convolution approaches.

These recent advances beg the question of why not apply sparse methods
on Winograd transform. The potential gain of this combination is
obvious. To realize this potential, however, one must
(1) be able to significantly prune away Winograd coefficients of
a CNN with minimal impact to the CNN's accuracy, and (2)
develop computation implementations that can exploit the pruned
Winograd parameters well enough that result in meaningful inference
speedups.
It turns out that pruning Winograd parameters is challenging
for reasons we will explain shortly. The Winograd sparsity
achieved reported so far is only moderate, which may explain why
there is no published effort on optimized sparse Winograd convolution implementations
to boost inference speed. This paper reports advances in both fronts,
illustrated by pruning Winograd kernels to more than 90\% sparsity while containing accuracy loss to within
0.1\%; as well as an actual up to 5.4$\times$ and 2.1$\times$ speedups of our sparse Winograd convolution compared to dense direct and dense Winograd convolution, respectively.

Pruning the original
convolution coefficients (the ``spatial'' domain) does not in general
result in sparse Winograd kernels. More fundamentally,
the linear transform that maps spatial to Winograd
parameters is non-invertible as there are more Winograd than spatial
coefficients. Thus any training and pruning method that needs
to make use of both sets of parameters in conjunction will cause inconsistency, which in turn leads to major accuracy loss when achieving acceptable sparsity.

We tackle this challenge and show that pruning Winograd parameters becomes successful
when we replace the convolution layer in question by
a {\it Winograd layer}, eliminating the need to use both the spatial
and Winograd parameters in conjunction. Our effective pruning method then paves the road to our highly optimized sparse Winograd convolution to materialize the sparsity for high speed inference. Figure~\ref{fig:architecture-explained} illustrates the architecture
of a Winograd layer using a simple 1D convolution of a 2-length spatial kernel
with 4-length input. The corresponding Winograd kernel is 3-length.
Replacing a convolution layer by a Winograd layer has several advantages. First, from a conceptual point of
view, this new layer architecture reflects directly the relationship between the key parameters
of the computation (which are the Winograd parameters) and the overall neural network.
Second, as alluded to earlier,
we have more Winograd than spatial parameters. Thus a Winograd layer in fact
has a larger capacity as we are no longer restricted to use only those Winograd
parameters that actually correspond to a set of spatial convolution parameters.
Last but most important, steering of the training process such as pruning becomes
possible and straightforward. We emphasize that this is not the case if one
employs the spatial convolution layer:
the non-invertible mapping between convolution and
Winograd parameters is a major obstacle. While this obstacle may be overcome via approximation for small networks (LeNet) and datasets (MNIST) as reported in~\cite{sparse_winograd}, our experiments (Section~\ref{subsec:sparsity_accuracy}) show that approximation does not work for larger networks (AlexNet) and datasets (ImageNet). 

The main contributions of this paper are as follows:
\begin{enumerate}[leftmargin=*, noitemsep, topsep=-10pt]
\item We define and
advocate the use of the Winograd layer as an architecture. Details of
the forward and backward passes of this layer are derived and training
methods aim to prune are devised
(Sections~\ref{sec:winograd_layer_new}--\ref{sec:winograd_layer_sparsify}).
\item We design and implement a highly optimized sparse Winograd convolution computation for fast inference by exploiting high sparsity obtained by our pruning methods (Section~\ref{sec:sparse_winograd_convolution}).
\item We demonstrate the effectiveness of training and pruning with
the Winograd-layer architecture (Section~\ref{sec:results}) and fast inference with our sparse Winograd convolution. Particularly, we prune AlexNet~\citep{alexnet} to eliminate its Winograd parameters by more than 90\%, while maintaining its original accuracy.
This leads to up to 2.1$\times$ speedups over an ideal dense Winograd convolution, one of the fastest convolution algorithms to date~\citep{winograd}.
\end{enumerate}

\section{Related Work}\label{sec:related_works}

That convolution can be computed with less arithmetic complexity through the use
of transformation is well understood. \cite{fft-cnn,fbfft} appear to be the
first ones that detailed the use of FFT as a compute kernel in Deep Learning. This kernel
executes the ``spatial-domain'' convolution by element-wise multiplication
in the ``frequency-domain''.
More recently, \cite{winograd} shows convincingly that Winograd transforms
outperform FFTs in the common convolution use cases in Deep Learning. These works
illustrate the use of transform methods (FFT or Winograd) as computation kernels
to speed up a normal inference or training process.

Recently, there have been a few research attempts to further reduce compute requirements and memory footprint of Winograd convolution by pruning. However, they have had limited success in pruning and not shown actual speedups in inference. \cite{sparse_winograd} attempts to prune parameters in the
transformed domain. For Winograd parameters, the overall network's architecture is
that of the original CNN, but the forward pass is performed using the Winograd kernel
with some of the parameters set to zero. A backward pass is performed that updates
the original ``spatial-domain'' parameters. We believe there is an inconsistency
in this model caused by the fact that mapping between spatial-domain and Winograd convolution parameters
is non-invertible. In general, there is no spatial-domain convolution parameters that correspond to
the modified (masked off) Winograd parameters that are used to compute all the
forward-pass intermediate quantities, while the backward pass computes the
gradient using these intermediate quantities \emph{in conjunction with} the
spatial-convolution-kernel parameters. While they achieve reasonable results~\cite{sparse_winograd} with LeNet~\citep{lecun1998gradient} on the MNIST
dataset~\citep{lecun1998mnist}, 
our results on larger networks and datasets such as AlexNet~\citep{alexnet} on
ImageNet~\citep{imagenet} as shown in Section~\ref{subsec:sparsity_accuracy} demonstrate that significant accuracy loss and/or low sparsity are inevitable for such approaches as in~\cite{sparse_winograd}. 
Indeed, direct pruning of the parameters in our proposed Winograd layers
is how we finally overcome this accuracy problem. 

\cite{sparse_winograd2}, a concurrent work to ours, similarly addresses the non-invertible issue by
directly pruning in Winograd domain.
Interestingly, they also move the ReLU operation into Winograd domain to obtain sparsity in
activations as well.
However, their sparsity is lower than ours (75\% vs. 90+\%) and is
evaluated with smaller dataset (CIFAR-10 vs. ImageNet).
Moreover, their work does not show actual inference speedups from sparsity in Winograd domain.
Our work provides a highly optimized sparse Winograd convolution design and implementation for fast inference.

In a conceptual sense, \cite{spectral-cnn} relates
closely to our current work. The authors advocate representing
the convolution kernels in frequency domain and detailed the gradient calculation with respect
to the frequency parameters. The network architecture, however, remains in the original form as
outlined in Section 4 of the paper: When convolution is to be computed, the frequency parameters
must first be transformed back to spatial domain in which a regular convolution is performed.

Since pruning in spatial domain does not provide a high enough sparsity in Winograd parameters to benefit from
general sparse representations such as compressed sparse row (CSR), a hardware feature for zero-skipping has been proposed to take advantage of the low sparsity in Winograd parameters~\citep{park2016zero}. Our paper shows that, by directly pruning in Winograd domain, we can obtain high enough sparsity to speedup inference without such specialized hardware features.

\section{Winograd Layers --
Definition and Backpropagation} \label{sec:winograd_layer_new}

Consider a typical convolution layer where the input tensors with $C$ channels
of features maps each of dimension $H_i\times W_i$ are transformed into $K$
output channels via a simple unit-stride, unit-dilation linear
convolution with kernels of size $r\times s$:
\begin{equation}\label{eqn:conv-std}
\begin{split}
\tens{I}\in\reals^{C\times H_i \times W_i} \rightarrow &
\tens{O}\in\reals^{K\times H_o \times W_o} \\
\quad {\rm via}
\quad \tens{O}(k,:,:) = & \sum_{c=1}^C \tens{W}(k,c,:,:) \conv \tens{I}(c,:,:)
\end{split}
\end{equation}
where $H_o = H_i-r+1$ and $W_o=W_i-s+1$ and $\conv$ stands for 2D linear convolution.


The computation of Equation~\ref{eqn:conv-std} can be performed using the
Winograd transform which has a lower arithmetic complexity than Equation~\ref{eqn:conv-std}
suggests. The details of this computation
we present now are crucial to our definition of {\it Winograd layers} and their training.
A convolution $\tens{W}(k,c,:,:)\conv\tens{I}(c,:,:)$ with $\tens{I}(c,:,:)$
of size $H_i\times W_i$ can be broken down into many convolutions each involving
smaller tiles of $\tens{I}$.
We illustrate this ``overlapping'' method by the following one-dimensional example,
using self-explanatory Matlab-like array index notation.
\[
\begin{split}
&  W(0:2) \conv I(0:5) \rightarrow O(0:3)\\
& \quad {\rm by} \quad
 W(0:2) \conv
\left[\begin{array}{c} I(0:3) \\ I(2:5) \end{array}\right] \rightarrow
\left[\begin{array}{c} O(0:1) \\ O(2:3) \end{array}\right]
\end{split}
\]
Note that $I$ is broken up into two tiles with some duplicating elements while $O$ is partitioned
(without duplication) into two tiles. More generally, given convolution kernels of size $r\times s$
and (small) sizes $m, n$ that divide\footnote{
This assumption simplifies the presentation and can be easily eliminated by for example
zero padding.}
$H_o$ and $W_o$, respectively,
we reshape the input and output tensors $\tens{I}$, $\tens{O}$ into $\tilde{\tens{I}}$, $\tilde{\tens{O}}$
\[
\begin{split}
\tens{I}\in\reals^{C\times H_i\times W_i} \rightarrow
\tilde{\tens{I}}\in\reals^{C\times T\times (m+r-1)\times (n+s-1)}\\
\quad {\rm and} \quad
\tens{O}\in\reals^{K\times H_o\times W_o} \rightarrow
\tilde{\tens{O}}\in\reals^{K\times T\times m\times n} .
\end{split}
\]
The value $T$ is the number of resulting tiles of the reshaping,
$T = H_o\,W_o/(mn)$. The input tile size is $(m+r-1)\times (n+s-1)$, while the output tile size is $m\times n$.
We express the reshaping by two index mapping functions $\phi$ and $\psi$
\[
\tilde{\tens{I}}(c,t,i,j) = \tens{I}(c,\phi(t,i,j)), \quad
\tens{O}(k,i,j) = \tilde{\tens{O}}(k,\psi(i,j))
\]
where $\phi$ is many-to-one and maps a 3-tuple to a 2-tuple while
$\psi$ is invertible and maps a 2-tuple to a 3-tuple.
Using the overlapped form, we have
\begin{equation}\label{eqn:conv-overlapped}
\tilde{\tens{O}}(k,t,:,:) =
\sum_{c=1}^C \tens{W}(k,c,:,:) \conv \tilde{\tens{I}}(c,t,:,:).
\end{equation}

Computing Equation~\ref{eqn:conv-overlapped} with
straightforward convolution takes $mnrs$ multiplications in each of the summands.
In contrast, Winograd's method can possibly need only as few as
$(m+r-1)(n+s-1)$ multiplications via:
\begin{equation}\label{eqn:conv-winograd}
\begin{split}
&\tilde{\tens{O}}(k,t,:,:)  = A_1^T \\
& \times \left[
\sum_{c=1}^C
\left(
G_1\tens{W}(k,c,:,:)G_2^T
\right)
\odot
\left(
B_1^T\tilde{\tens{I}}(c,t,:,:)B_2
\right)
\right]
A_2.
\end{split}
\end{equation}
The six matrices $A_j, G_j, B_j$, $j=1,2$, (of consistent dimensions)
are independent of $\tens{W}$ and $\tilde{\tens{I}}$, and $\odot$ is element-wise multiplication. In
many instances the $A_j, B_j, C_j$ matrices are so simple that applying them requires
no multiplications. For example when $r=s=3$ and $m=n=2$, $A_1=A_2$, $B_1=B_2$
and $G_1=G_2$ are simple matrices with $0, \pm 1$, and $\pm 1/2$ are entries
(see~\cite{winograd} for example).

Motivated by this, we define a Winograd layer to be a topology
specified by a tensor
$\tens{W}_F \in \reals^{K\times C\times (m+r-1)\times (n+s-1)}$
that computes $\tens{O}$ from $\tens{I}$ via
\begin{align}\label{eqn:winograd_layer}
\begin{split}
&\tilde{\tens{I}}(c,t,i,j) = \tens{I}(c,\phi(t,i,j)), \\
&\tilde{\tens{O}}(k,t,:,:) \\
&= A_1^T
\left[
\sum_{c=1}^C
\left(
\tens{W}_F(k,c,:,:)
\right)
\odot
\left(
B_1^T\tilde{\tens{I}}(c,t,:,:)B_2
\right)
\right]
A_2, \\
&\tens{O}(k,i,j) = \tilde{\tens{O}}(k,\psi(i,j))
\end{split}
\end{align}
Since $m, n >1$ in practice, $(m+r-1)(n+s-1) > rs$ and a
Winograd layer (Equation~\ref{eqn:winograd_layer}) has a higher
capacity than a corresponding convolution layer (Equation~\ref{eqn:conv-std}).



To incorporate a Winograd layer within a standard CNN framework (e.g., Caffe) so as to
allow training and inference, it suffices to be able to compute the forward and backward passes.
The forward pass is straightforward as it simply follows
Equation~\ref{eqn:winograd_layer}, for which we note that~\cite{winograd}
details an optimized implementation.
For the backward pass, we need to compute the partial derivatives of the
scalar loss function $L$
w.r.t. each of the variables $\tens{I}(c,i,j)$ and $\tens{W}_F(k,c,i,j)$ in terms of the
known partial derivatives of $L$ w.r.t. $\tens{O}(k,i,j)$, or $\partial L/\partial \tens{O}$
in short. We present the derivations here as they pertain to the Winograd layers
and are thus unavailable elsewhere.\footnote{Note that the Winograd-kernel approach~\citep{winograd}
does not consider Winograd as a layer architecture with full fledged backward propagation.
They use Winograd solely to accelerate the convolution operations in both
forward and backward propagation while all parameters reside in the spatial domain.}

First, we use this key form of chain rule: Suppose the partial derivatives of a scalar function
$L$ w.r.t. an array of variables $y_{ij}$, $Y\in\reals^{\mu\times\nu}$, are known. Moreover,
the variables $Y$ are in fact dependent variables of an array of variables $x_{ij}$,
$X\in\reals^{\mu'\times\nu'}$ via $Y = U^T X V$ where $U, V$ are constant matrices
of commensurate dimensions. The partial derivatives of $L$ w.r.t. $X$ are then given by
$\partial L/\partial X =
U\;  \partial L/\partial Y \; V^T.$

Denote the intermediate variables in Equation~\ref{eqn:winograd_layer} by
$\tilde{\tens{I}}_F$ and $\tens{Z}$:
$\tilde{\tens{I}}_F(c,t,:,:) = B_1^T\,\tilde{\tens{I}}(c,t,:,:)\,B_2$ and
$\tens{Z}(k,t,:,:) = \sum_{c=1}^C \tens{W}_F(k,c,:,:)\odot\tilde{\tens{I}}_F(c,t,:,:)$
for all applicable indices $c,k,t$. Note
in particular that
$\tens{Z}(:,:,i,j) = \tens{W}_F(:,:,i,j) \tilde{\tens{I}}_F(:,:,i,j)$, that is the 2-dimensional slice
of $\tens{Z}$ with any fixed $i,j$ is a matrix product.
We can then compute $\partial L/\partial \tens{W}_F$
using the Chain Rule via
\begin{align}\label{eqn:partial_L_partial_WF}
\begin{split}
\frac{\partial L}{\partial \tens{Z}(k,t,:.:)}
 = &
A_1\,\frac{\partial L}{\partial \tilde{\tens{O}}(k,t,:,:)}\,A_2^T
\\
\quad
\frac{\partial L}{\partial \tens{W}_F(:,:,i,j)}
 = &
\frac{\partial L}{\partial \tens{Z}(:,:,i,j)}\,
\left(\tilde{\tens{I}}_F(:,:,i,j)\right)^T
\end{split}
\end{align}
noting that $\partial L/\partial \tilde{\tens{O}}$ is
$\partial L/\partial \tilde{\tens{O}}$ with a simple index mapping $\psi^{-1}$.

Similarly, we use the Chain Rule to obtain $\partial L/\partial \tens{I}$, using $\partial L/\partial \tens{Z}$ computed above:
\begin{align}\label{eqn:partial_L_partial_I}
\begin{split}
\frac{\partial L}{\partial \tilde{\tens{I}}_F(:,:,i,j)}
 = &
\left(\tens{W}_F(:,:,i,j)\right)^T\,\frac{\partial L}{\partial \tens{Z}(:,:,i,j)}
\\
\quad
\frac{\partial L}{\partial \tilde{I}(c,t,:,:)}
 = &
B_1\,\frac{\partial L}{\partial \tilde{\tens{I}}_F(c,t,:,:)}\,B_2^T 
\\
\quad
\frac{\partial L}{\partial \tens{I}(c,i,j)}
 = &
\sum_{\substack{t,i',j'\;\text{where} \\\ (i,j) = \phi(t,i',j')}}
\frac{\partial L}{\partial \tilde{\tens{I}}(c,t,i',j')}
\end{split}
\end{align}

In summary, the backward propagation of Winograd layers is implemented by Equations~\ref{eqn:partial_L_partial_WF}
and~\ref{eqn:partial_L_partial_I}.

\section{Winograd Layers -- Training and Pruning}
\label{sec:winograd_layer_sparsify}

Consider a $L$-layer network with a mixture of layers such as
convolution, fully connected, and pooling.
Let $\bsigma = [\mathbf{v}_{(1)},\mathbf{v}_{(2)},\ldots,\mathbf{v}_{(L)}]$
be the parameters in the usual spatial domain, and 
$\btheta = [\mathbf{w}_{(1)},\mathbf{w}_{(2)},\ldots,\mathbf{w}_{(L)}]$
be the parameters when the convolution layers are replaced by Winograd layers.
Let ${\cal L}_s(\bsigma)$ and ${\cal L}_w(\btheta)$ be the respective loss
functions. Training is typically done by minimizing an
energy function that is the sum of the loss and a regularization penalty.
We aim to arrive at a $\btheta^*$ such that many of its elements, including
those within the Winograd layer, are zero.
Pre-training, pruning, and fine-tuning (re-training) are the three major steps
to achieve this goal.

\subsection{Pre-training in spatial domain}
In pre-training, we try to obtain a parameter $\btheta^{({\rm start})}$ that makes
the network accurate but is also amenable to pruning in the next stage. Because many
open-source networks provide already-trained weights (in the spatial domain), the pre-training method we adopt here starts with these weights. We apply
the standard SGD algorithm on the energy function of the form 
$E(\bsigma) = {\cal L}_s(\bsigma) + \lambda\,R(G(\bsigma))$.
Here $R$ is a regularization function applied on $G(\bsigma)$, which converts ``on-the-fly''
the parameters related to the convolution layers to the Winograd coefficients using
the $G_j$ matrices shown in Equation~\ref{eqn:conv-winograd}, in order to encourage sparsity in Winograd domain.
We note that common frameworks such as Caffe allow the incorporation of
various regularization penalties on different layers of a network in a 
straightforward manner.
At the end of the SGD iterations, the obtained spatial parameter $\bsigma^*$ is mapped into the Winograd domain
by the same function $G$: $\btheta^{({\rm start})} \leftarrow G(\bsigma^*)$.

\subsection{Pruning with Regularization and Gradient-based Thresholding}
\label{pruning}
Pruning is done directly in Winograd domain, which is enabled by the Winograd layer. Regularization and gradient-based thresholding are the two techniques used for pruning. Regularization is a common and useful technique to attenuate over-fitting and induce sparsity.
The energy function is
\begin{equation}\label{eqn:energy_function}
E(\btheta) = {\cal L}_w(\btheta) + R(\btheta),
\qquad
R(\btheta) = \sum_{\ell=1}^L \lambda_\ell \| \mathbf{w}_{(\ell)}\|_{p_\ell}.
\end{equation}
Common choices of norms are $p_{\ell}=1$ or $2$, and we use $L1$-norm for the layers to be pruned.

To achieve higher sparsity, thresholding is used in addition to regularization.
The idea of thresholding is to set to zero particular
entries within the parameter $\btheta^{(k)}$ that are deemed inconsequential~\citep{prune,sscnn,dns}.
Instead of using one uniform threshold to judge the significance, the threshold
on a particular parameter is dependent on how significantly this parameter affects
the underlying loss function $L$. The thresholding function
$T_{\epsilon,\beta}$ is of the form
\begin{equation}\label{eqn:thresholding_function}
T_{\epsilon,\beta}( w ) =
\left\{
\begin{array}{l l}
0 & \mbox{if $|w|\left(|\frac{\partial L}{\partial w}(\btheta)|+\beta \right) < \epsilon$} \\
w & \mbox{otherwise}
\end{array}.
\right.
\end{equation}
We apply the thresholding function to a vector simply by applying it
on each element, denoted with the slightly abused notation: $T_{\epsilon,\beta}(\mathbf{w})$.
The numbers $\epsilon$ and $\beta$ are part of the ``hyper-parameters'' of the training procedure.
This scheme uses smaller thresholds as the magnitude of gradient increases (threshold is $\epsilon/\beta$, $\epsilon$, and 0, when the magnitude
of gradient is 0, $1 - \beta$, and $\infty$, respectively).
We find that the choice of $\epsilon$=1e-4 and $\beta$=0.1 works well.
We emphasize that thresholding the parameters including Winograd parameters is now straightforward
because of the Winograd parameters are the direct independent parameters of the Winograd layers.

In the Winograd domain, where the L-layer network contains both Winograd layers and other layers (e.g., pooling) with parameters: $\btheta = [\mathbf{w}_{(1)},\ldots,\mathbf{w}_{(L)}]$, the pruning step applies regularization and thresholding as follows where
$\nabla_{\cal B}$ is the familiar stochastic gradient with a
mini batch ${\cal B}$ and $\eta_k$ is the learning rate:
\begin{align}\label{eqn:winograd_training_1}
\begin{split}
E(\btheta)
= & {\cal L}_w(\btheta) + \sum_{\ell=1}^L \lambda_\ell \|\mathbf{w}_{(\ell)}\|_{p_\ell}
\\ 
\quad
\btheta^{(k+1)}
\leftarrow &
T_{\epsilon,\beta} \left( 
\btheta^{(k)} - \eta_k \nabla_{\cal B} E(\btheta^{(k)}) \right)
\end{split}
\end{align}



\subsection{Fine-Tuning: Recover Accuracy Loss}
\label{fine_tuning}
Similar to pruning in spatial domain, pruning Winograd parameters will cause accuracy loss,
which necessitates a fine-tuning step to recover the loss.
Same as pruning, fine-tuning is only done in the Winograd domain. During fine-tuning, the zero parameters obtained from the pruning step are fixed, while the network is trained to adjust the other non-zeros parameters to recover accuracy loss. We use L2 regularization during fine-tuning.
The larger capacity of Winograd domain gives another benefit here. Even with high sparsity, the remaining degrees of freedom allow a better recovery
of accuracy by the fine-tuning step. This further allows in general
more aggressive regularization and thresholding during the pruning step.

\section{Speeding Up Inference with Sparse Winograd Convolution}
\label{sec:sparse_winograd_convolution}

A highly optimized sparse Winograd Convolution implementation is paramount
to
turn the sparsity obtained into actual performance gain in inference
computations.
Winograd convolution consists of three steps: (1) input transformation that
corresponds to multiplications of each tile with small matrices $B_1$ and $B_2$
in Equation~\ref{eqn:conv-winograd}, (2) element-wise multiplications in
Winograd domain, and (3) output inverse transformation that corresponds to
multiplications with $A_1$ and $A_2$.
The bulk of computation is in the second step, which can be implemented as
$(m+r-1)(n+s-1)$ independent multiplications of $K$$\times$$C$ matrices with
$C$$\times$$T$ matrices, where we have number of $T$ tiles of $m$$\times$$n$ in size, $r$$\times$$s$ sized
convolution kernels, $C$ input channels, and $K$ output channels.
With a model pruned in Winograd domain, the $K\times C$ matrices are sparse and
the matrix multiplications become sparse-matrix times dense-matrix
multiplications (SpMDM).

We first parallelize all three steps over multiple images within a batch.
When the batch is not large enough for each thread to have its own
image, we further parallelize over input/output channels.
This minimizes data exchanges among cores and also allows fusing the second and
third steps for cache locality.
Alternatively, the second step can be parallelized over $(m+r-1)(n+s-1)$
independent matrix multiplications.
This approach has a benefit of increasing the multiplication size to
$K$$\times$$C$ by $C$$\times$$(N\cdot T)$, where $N$ is the number of images per batch,
exploiting more reuse out of the $K$$\times$$C$ matrices.
However, this approach varies decomposition over the three steps,
incurring significant data exchanges among the cores.
Our scheme replicates the $K$$\times$$C$ matrices at each core,
but this is mitigated by that the $K$$\times$$C$ matrices are sparse and
can be much smaller to fit L1 or L2 caches.

Libraries for dense convolution operations in CNN often layout the data
interleaving multiple channels so that vectorization can be done over channels.
However, since the sparsity pattern of parameters varies over channels, this layout and
vectorization scheme is inefficient for SpMDM in the second step of sparse
Winograd convolution.
Therefore, we use a layout where the fastest moving dimension is tiles and
vectorize over tiles.
Our open source project will show more implementation details (link omitted for
double blind review).


\section{Experiments and Results}\label{sec:results}

This section describes Winograd training/pruning and sparse Winograd inference results. We implement Winograd layer (forward and backward propagation for training/pruning) as in Section~\ref{sec:winograd_layer_new} and sparse Winograd convolution as in Section~\ref{sec:sparse_winograd_convolution} in our branch of Caffe~\citep{caffe}.

\subsection{Winograd Convolution Pruning: Sparsity and Accuracy}


We use pre-trained spatial domain model to save overall training time. Particularly, we start with the Caffe reference model from the Caffe model zoo (we call it AlexNet for simplicity even though it is a slight variation).
We use the ImageNet ILSVRC-2012 dataset for pruning and test.
Since the first convolution layer does not provide high enough sparsity to get speedups~\citep{prune,sscnn}, we do not attempt to prune that layer.
We use the gradient-based thresholding described in Section~\ref{sec:winograd_layer_sparsify} with $\epsilon$=1e-4 and $\beta$=0.1.
We use learning rates of 5e-5 and 1e-4, and regularization factors of 5e-4 and 5e-5 in the pruning and fine-tuning steps, respectively. %
We use 200$\times$ smaller learning rates for the Winograd layers
to ensure convergence. 


\label{subsec:sparsity_accuracy}

Table~\ref{tab:results} lists the accuracy and the sparsity of {\tt conv2-5} layers in AlexNet.
Our method of training and pruning directly in Winograd domain (method A)
results in 90.6--95.8\% sparsity with only 0.1\% top-1 accuracy drop from the
reference model.
Method B maintains convolution parameters both in spatial and Winograd domains
and applies thresholding in 3 steps: (1) temporarily transform
spatial parameters to Winograd, (2) threshold the Winograd parameters, and (3)
find the least-square projection that maps the parameters back to the spatial
domain.
Since we have more parameters in Winograd domain, the Winograd parameters cannot be
inversely transformed to the spatial domain exactly, hence the least-square projection.
Due to this non-invertibility, method B either drops the accuracy by 8.1\% (B1
in Table~\ref{tab:results}) or results in much lower sparsity (B2).
Method C is from recent spatial domain pruning results~\citep{iclr17}, which
shows that, even when a model has $\sim$90\% high sparsity in spatial domain,
the sparsity significantly degrades to 25--70\% once converted to Winograd
domain.

The results shown in Table~\ref{tab:results} use gradient-based thresholding described in Section~\ref{sec:winograd_layer_sparsify}, where
atop regularization the gradient-based thresholding further reduces the non-zero density of each layer by up to 1.3$\times$
without affecting accuracy.
The fine-tuning step improves the accuracy from 56.5\% to 57.3\% in method A.
When natively pruned in Winograd domain (method A), we frequently find sparsity
patterns that have no counter parts in spatial domain such as a 6$\times$6
kernel in Winograd domain with only one non-zero at the second row of the
second column.
This illustrates that CNN architectures with layers of directly trainable parameters
in Wingorad domain are more expressive and have more opportunity for pruning.

\begin{table*}[t]
\caption{\small The top-1 test accuracy and sparsity of 3$\times$3 and 5$\times$5 convolution layers of AlexNet resulted from different pruning methods. The accuracy of the original AlexNet is 57.4\%.
Method {\tt B} uses spatial and Winograd parameters in conjunction similarly to \cite{sparse_winograd}.
{\tt B1} and {\tt B2} are with different hyper-parameters, with {\tt B1} to match the sparsity of method A and {\tt B2} to match the accuracy.
For method {\tt C} \citep{iclr17}, Winograd domain sparsity is obtained by transferring spatial domain kernels to Winograd domain kernels, with initial spatial domain sparsity shown inside parenthesis.
}
\label{tab:results}
\centering
\small
\ \\
\begin{tabular}{c | c | c c c c }
\hline
                         & Top-1    & \multicolumn{4}{c}{Sparsity of Convolution Layers in Winograd domain} \\
\multirow{-2}{*}{Method} & Accuracy & {\tt conv2} & {\tt 3} & {\tt 4} & {\tt 5}           \\
\hline
A (Ours)                 & 57.3\%   & 90.6\%      & 95.8\%  & 94.3\%  & 93.9\%            \\
\hline
B1                       & 49.3\%   & 92.8\%      & 94.2\%  & 93.2\%  & 91.1\%             \\
B2                       & 57.3\%   & 54.1\%      & 67.5\%  & 62.4\%  & 60.2\%             \\
\hline
C \citep{iclr17}         & 57.4\% & 25.3\% (85.6\%) & 69.3\% (93.1\%) & 66.0\% (91.8\%) & 61.3\% (88.5\%) \\
\hline
\end{tabular}
\end{table*}

\begin{figure*}[t]
\includegraphics[width=1.02\textwidth]{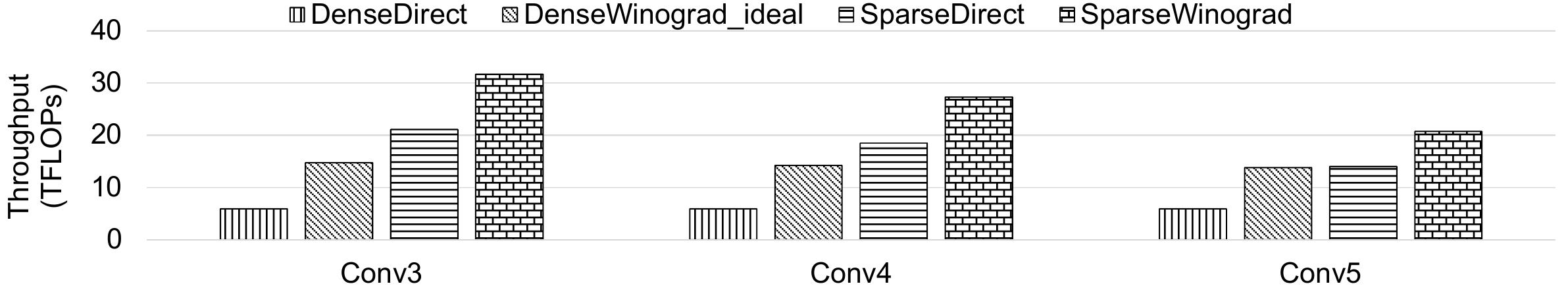}%
\caption{\small
Performance of sparse Winograd convolution compared with other methods,
evaluated with 3$\times$3 convolution layers of pruned AlexNet model shown in
Table~\ref{tab:results} and batch size 56 (i.e., 1 image per core).
Except dense direct convolution, performance is evaluated as effective FLOP/s
computed by (the number of floating-point operations that would have been
executed by dense direct convolution)/(execution time).
}
\label{fig:sparse_winograd_conv}
\end{figure*}

\subsection{Sparse Winograd Convolution: Speedup Inference}

Our sparse Winograd convolution inference is evaluated on a dual-socket server
with Intel Xeon Platinum 8180 processors running at 2.5GHz, with total 56 cores
and 77 MB last-level cache. This platform represents the latest
server-class systems inside data centers.
In Figure~\ref{fig:sparse_winograd_conv}, {\tt SparseWinograd} shows
layer-wise performance of our sparse Winograd convolution design with the our
obtain sparsity shown as method A in Table~\ref{tab:results}.
{\tt DenseDirect} is measured with libxsmm~\citep{libxsmm}, a
state-of-the-art open source dense convolution implementation.
{\tt DenseWinograd\_ideal} is an ideally projected performance assuming that
the speedup over {\tt DenseDirect} is commensurate with the reduction in
floating-point operations by using Winograd algorithm.
We use this projection because there has yet to be a dense Winograd
implementation optimized enough for the evaluated platform.
Note that the ideal speedup is not usually realizable because Winograd
convolution is less compute intensive and its performance is more bandwidth
bound than direct convolution.
{\tt SparseDirect} is measured with an open source sparse direct
convolution implementation from \cite{iclr17} with the model pruned in spatial
domain shown as method C in Table~\ref{tab:results}.
{\tt SparseWinograd} constantly outperforms the other methods: up to
5.4$\times$ over {\tt DenseDirect}, 2.1$\times$ over {\tt
DenseWinograd\_ideal}, and 1.5$\times$ over {\tt SparseDirect}. Since dense Winograd convolution has been demonstrated to be the fastest~\cite{winograd} for small convolution kernels in popular CNNs, the 2.1$\times$ speedup of our spare Winograd over the ideal dense Winograd shows its great potential in accelerating inference for popular CNNs.

\section{Conclusion}
\label{sec:conclusion}
As CNN has become pervasive, the relentless pursuit of fast convolution has inspired new algorithms and techniques for accelerating convolution. Transformation methods, especially Winograd convolution, and sparse methods are among the most successful approaches.
This paper is the first to successfully combine them to construct sparse Winograd convolution. Moreover, we have demonstrated that our sparse Winograd can achieve 90+\% sparsity without accuracy loss, leading to more than 5.4$\times$ speedup over dense direct convolution in 3$\times$3 convolution layers of AlexNet on a latest Intel Xeon CPU.
Looking ahead, our next step includes application to deeper networks like GoogLeNet~\citep{googlenet} and ResNet~\citep{resnet} and to other platforms like FPGAs.


\if{false}
\section{Some other thoughts}
First, on the derivatives with respect to time-domain weights and Winograd weights.
Don't know if the following matches Jongsoo's
experiments, but here is my observation. Suppose we do the following.
\newcommand{\paramtime}{\boldsymbol{\gamma}_{T}}
\newcommand{\paramwinograd}{\boldsymbol{\gamma}_{W}}
\begin{enumerate}
\item Initialize the time domain network and use the same
initialization for the winograd domain training -- that is, the winograd parameters are obtained by transforming the time-domain
parameters. And we only use winograd on the first convolutional layer.
\item One forward pass on both networks. Thus obtaining $\tens{O}, \tens{O'}, L_{\rm time}, L_{\rm win}$. $\tens{O}$ and $\tens{O'}$
should be the same, as are the two loss function values.
\item One backward pass on both networks, let $\paramtime$ be the time-domain parameters of the first layer and
$\paramwinograd$ be the winograd. After the first backward pass, we will have
$\frac{\partial L}{\partial \paramtime}$ and
$\frac{\partial L}{\partial \paramwinograd}$.
Because
$$
\paramwinograd = \tens{M}\, \paramtime
$$
(where $\tens{M}$ is tall-skinny), applying the chain rule gives
\begin{equation}\label{eqn:paramtime-winograd}
\frac{\partial L}{\partial \paramtime} = \tens{M}^T\,\frac{\partial L}{\partial \paramwinograd}
\end{equation}
Thus computes $\tens{M}^T\,\frac{\partial L}{\partial \paramwinograd}$ using $\frac{\partial L}{\partial \paramwinograd}$
from the winograd-domain network, these values should match $\frac{\partial L}{\partial \paramtime}$ obtained in the time-domain
network.
\end{enumerate}

This match will not last to the second iteration because updating the time-domain and winograd domain parameters independently
will destroy the relationship $\paramwinograd = \tens{M}\, \paramtime$. Another observation, because $\tens{M}^T$ is short and fat,
it is possible to have large-norm $\paramwinograd = \tens{M}\, \paramtime$ while the norm of $\frac{\partial L}{\partial \paramtime}$
remains moderate.

The update steps will not preserve the initial relationship between time-domain and winograd-domain parameters.
\begin{eqnarray*}
\paramwinograd' & = & \paramwinograd - \delta_{w} \frac{\partial L}{\partial \paramwinograd} \\
\paramtime'     & = & \paramtime     - \delta_{t} \frac{\partial L}{\partial \paramtime}
\end{eqnarray*}

In general, we may have too many winograd parameters. A couple of ideas:
\begin{itemize}
\item If training directly in winograd domain, one may want to impose a penalty to encourage $\paramwinograd = \tens{M}\, \paramtime$.
That is, $\paramwinograd$ lies in the range space of $\tens{M}$. This can be given as $\tens{N}^T \paramwinograd = 0$. That is,
$\paramwinograd$ is orthogonal to the normal to the range space of $\tens{M}$.
\item Or train in time domain with regularization on $\tens{M} \paramtime$. Use $\tens{M} \paramtime*$ to identify potential
sparsity of $\paramwinograd$. Retrain in winograd domain while setting a number of positions to be zero.
\end{itemize}
\fi

\bibliographystyle{icml2017}
\bibliography{main}
\end{document}